\documentclass[11pt]{article}

\usepackage[table,xcdraw]{xcolor}
\usepackage[]{acl}

\usepackage{times}
\usepackage{latexsym}
\usepackage{graphicx}
\usepackage{caption}
\usepackage{subcaption}
\usepackage{booktabs}
\usepackage[T1]{fontenc}

\usepackage[utf8]{inputenc}

\usepackage{microtype}

\title{Performance-Efficiency Trade-Offs \\ in Adapting Language Models to Text Classification Tasks}

\author{
    Laura Aina, Nikos Voskarides, Roi Blanco \\
    Amazon, Barcelona, Spain\\
    \texttt{\{eailaura,nvvoskar,roiblan\}@amazon.com}
}

\begin{document}
\maketitle
\begin{abstract}
Pre-trained language models (LMs) obtain state-of-the-art performance when adapted to text classification tasks.
However, when using such models in real-world applications, efficiency considerations are paramount.
In this paper, we study how different training procedures that adapt LMs to text classification perform, as we vary model and train set size.
More specifically, we compare standard fine-tuning, prompting, and knowledge distillation (KD) when the teacher was trained with either fine-tuning or prompting. 
Our findings suggest that even though fine-tuning and prompting work well to train large LMs on large train sets, there are more efficient alternatives that can reduce compute or data cost. 
Interestingly, we find that prompting combined with KD can reduce compute and data cost at the same time.


\end{abstract}

\section{Introduction}
State-of-the-art techniques in NLP, such as adapting pre-trained language models~(LMs) to downstream tasks, typically rely on large model and/or train set sizes~\citep{radford2018improving,devlin2019bert}.
In real-world applications, serving large models or having large train sets may be prohibitive due to budget constraints, too slow inference, or scarcity of expensive-to-obtain labeled data.
Thus, solutions to build models that save time, money, and energy are preferable~\citep{strubell-etal-2019-energy}. 
A powerful technique to reduce model size is knowledge distillation (KD)~\citep{hinton2015distilling}. 
KD requires the availability of a large unlabeled dataset --the \emph{transfer set}-- which is usually easy to gather for real-world applications since examples of task inputs are abundant. 
In KD, the transfer set is weakly labeled with the predictions of a \textit{teacher} model; then a smaller \textit{student} model is trained to match these soft labels. 
The goal is to retain the teacher's abilities in a more compact architecture.

In this work, we study how performance varies across different scenarios (i.e.,~model or train set size) when adapting pretrained LMs to downstream tasks. 
We focus on text classification as a prominent downstream task. 
We consider the following training procedures to adapt LMs for classification:
\begin{itemize}
\item \emph{finetuning (F)}: A classification layer is added to the LM architecture, and the parameters of that layer are jointly learnt with the rest of the LM parameters~\citep{devlin2019bert};
\item \emph{prompting (P)}: The LM is trained rephrasing the downstream task as word prediction following the objective the LM was trained on~\cite{liu2021pre}. This technique --also known as \textit{prompt-based fine-tuning}-- was found to outperform standard fine-tuning when using small train sets~\citep{schick2021true};
\item  \emph{finetuning + KD (F+KD)}: a LM is trained using fine-tuning (\emph{F}) and then KD is performed to obtain a smaller model~\cite{turc2019};
\item \emph{prompting + KD (P+KD)}: a LM is trained using prompting (\emph{P}) and then KD is performed to obtain a smaller model. We show that this training procedure can yield classifiers that are both compact and sample-efficient.\footnote{In analogous spirit to \emph{P+KD}, previous works used LM-generated data to train sentence embeddings~\citep{schick2021generating}, or to obtain compact commonsense models~\citep{west-etal-2022-symbolic}. Concurrent to our work, \citet{meng2022generating} propose to use data generated by large unidirectional models to transfer their zero-shot NLU abilities to more compact bidirectional models.}
\end{itemize}

We study how the aforementioned training procedures perform on multiple text classification datasets, as we vary model and data cost.
Model size is used as an estimate of model cost~\citep{dehghani2021efficiency}. 
While recent studies have focused on training cost~\citep{strubell-etal-2019-energy,kaplan2020scaling,izsak2021train,yao2021nlp,hoffmann2022training}, we instead focus on inference cost as the most concerning aspect of compute cost in real-world applications: when serving millions of users, inference happens very frequently and on large-scale.
On the other hand, train set size is used as a proxy for data cost, assuming the cost of annotating one example is a constant.  
This allows us to discuss sample efficiency, that is, the amount of data required to achieve acceptable performance. 

Our contributions are two-fold.
First, we show that \emph{P+KD} allows to reduce both model and train set size while retaining high performance.
Second, we extrapolate recommendations on how to efficiently adapt LMs for downstream tasks.
The trends we identify indicate that the cost of increasing model parameters or train set size is not always worth it: small models or models trained with little data often achieve comparable performance than larger models trained with more data.

\section{Experimental setup}
\label{sec:exp-setup}
\subsection{Models} We experiment with 5 BERT LMs of increasing size released by~\citet{turc2019}: BERT-mini, -small, -medium, -base, -large. %
These models have been trained on the same English text corpus, share the BERT architecture but differ in hidden size, number of hidden layers and attention heads. Therefore, they differ in \emph{inference speed}: for instance, BERT-mini is 7x faster than BERT-large. Please refer to Table~\ref{table:bert_models} for more details on the compared models and their relative inference speed.
\begin{center}
\begin{table}[t]
\centering
\small{
\begin{tabular}{lrr}
Model   & \# Parameters & Relative speedup \\  \midrule
BERT-large &  336.2M   & - \\
BERT-base &  110.1M  &   2.6 \\
BERT-medium  & 41.7M  & 5.1\\
BERT-small  & 29.1M   &  6.3 \\
BERT-mini  & 11.3M   & 7.7 \\ \bottomrule
\end{tabular}
\caption{Details of the BERT models used in our experiments. Relative speedup is measured with respect to BERT-large, based on average inference time of our final models (across training procedure, tasks and configurations) on a single 16GB GPU with batch size 32, except for BERT-large where we use a batch size of 8.}
\label{table:bert_models}}
\end{table}
\end{center}
\subsection{KD} We use \textit{pre-trained distillation}~\cite{turc2019}, where both the teacher and the student are pre-trained LMs that are adapted to the downstream task. 
This was shown to work better than training both models from scratch directly on the downstream task~\cite{turc2019}. 
When doing \emph{KD}, we use as teacher model the best on dev data among those for that train set size. 
For instance, for \emph{F+KD} on BERT-small with train set size 20, we use as teacher the BERT-large model trained with fine-tuning on 20 training examples. 
BERT-large is always used as the teacher model in KD; therefore it is never used as student model in \emph{F+KD} or \emph{P+KD}. 

\subsection{Prompting} 
We adapt a LM to the downstream task by tuning its weights to output the correct predictions on the train set, as in Pattern-Exploiting Training (PET)~\citep{schick2021exploiting,schick2021s}.
In its original formulation, PET trains multiple LM instances with different templates\footnote{A template is the way the task is set up. For instance, one template is to append ``All in all, it was...'' to the end of a review and map output adjectives to sentiment labels (e.g., “terrible” $\rightarrow$ 1; “great” $\rightarrow$5).} and then uses their predictions to obtain a single classifier from the original LM. 
For simplicity in experimentation and without substantial loss in accuracy, we instead follow~\citet{le2021many}: we tune the LM with a single template and use this as our final classifier.
Note that the original PET algorithm also involves KD, but differently from our setting, it is used to obtain a model of the \emph{same size} of the starting LM from the ensemble of LMs trained on different templates. 
In contrast, we distill \emph{a large LM to a smaller LM}.

\subsection{Datasets} We use 4 English text classification datasets~\citep{zhang2015character}: Yelp-full (sentiment; 5 classes);
Yelp-polarity (sentiment; 2 classes), 
Yahoo-questions (question; 10 classes); 
AG news (news article; 4 classes). The test size of each dataset is 50K, 38K, 60K, 7.6K, respectively.
These datasets are large enough to allow us to both flexibly explore the effect of train set size and also build a large transfer set of unlabeled examples to be used by KD. We sample 10K examples from the original train set of each dataset as the transfer set (discarding the gold labels of those examples). In preliminary experiments we varied the size of the transfer set from 5K to 10K without observing substantial differences in the trends; larger transfer sets can be explored in future work.
The aforementioned datasets were employed in the experiments of \citet{schick2021exploiting}; we build on their experimental setup and prompting templates.\footnote{The code to run our experiments was developed upon \citeauthor{schick2021exploiting}'s code: \url{https://github.com/timoschick/pet}}
For each task, we consider 11 exponentially growing train set sizes from 20 to 20480, sampled from the original train set.
We exclude from each train set a 10\% portion to be used for sampling dev sets, with a minimum of 20 examples.
Studies exploring train set size in analogous spirit to ours used either no dev set~\citep{schick2021exploiting,schick2021true} or one that is kept constant across train set size~\citep{le2021many}. 
We strike a balance between these approaches by using a dev set of proportional size to the train set, which is a realistic assumption.
Even if very small, having a dev set is useful as it gives an indication of quality during model development.
In all train and dev sets we balance the number of examples for each class. 

\subsection{Hyperparameter Search} We focus on batch size and learning rate, and on the task-specific prompting template~\citep{schick2021exploiting}.
Keeping these constant across experiments could be unfair, as different combinations of train sets sizes, models, training procedures and tasks may favor different hyperparameters.
On the other hand, to run a search for each combination would be extremely costly and time-consuming.
We go for an intermediate strategy; for instance, for BERT-mini trained with 80 examples, we use the hyperparameters selected for BERT-small with 20 examples.
We describe our choices in more detail in Appendix~\ref{sec:search}.

\subsection{Other Details} For every task and configuration (combination of LM and train set size; 5 $\times$ 11), we run training 4 times and report on the model achieving the highest accuracy on the dev set~\citep{le2021many}.
As there tends to be variation across runs~\citep{dodge2020fine,schick2021true}, we focus on the highest achieved accuracy, as opposed to the mean, in order to compare each configuration in its best case scenario. Note that we observe similar trends when considering mean and standard deviation across runs (see Fig.~\ref{fig:accuracy_mean} in the Appendix).

\section{Results and Discussion}

\begin{figure}[t]
    \centering
    \begin{subfigure}[t]{0.45\columnwidth}
        \centering
        \includegraphics[height=1.3in]{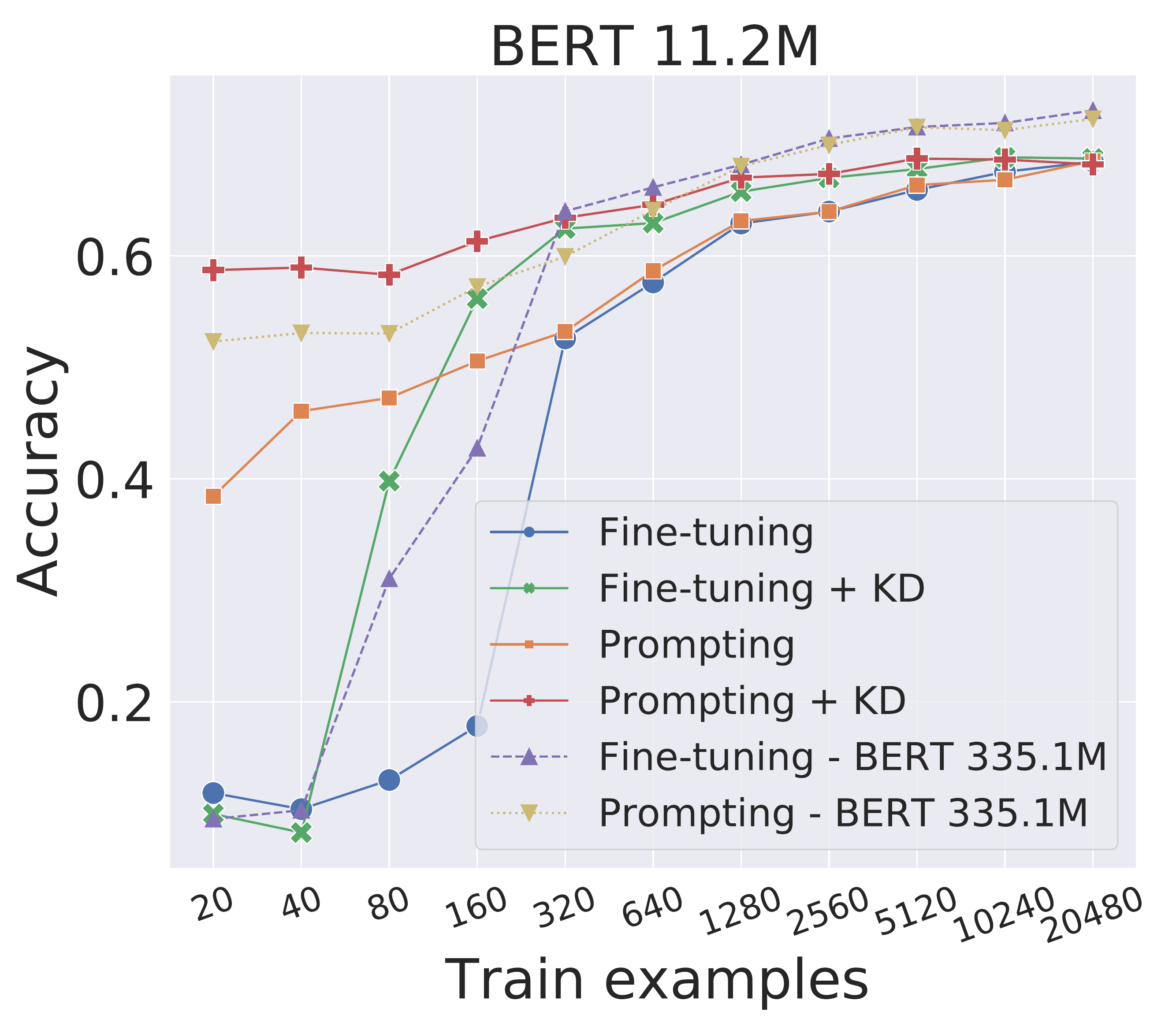}

        \caption{Yahoo-Q}
    \end{subfigure}
    ~
    \begin{subfigure}[t]{0.45\columnwidth}
        \centering
        \includegraphics[height=1.3in]{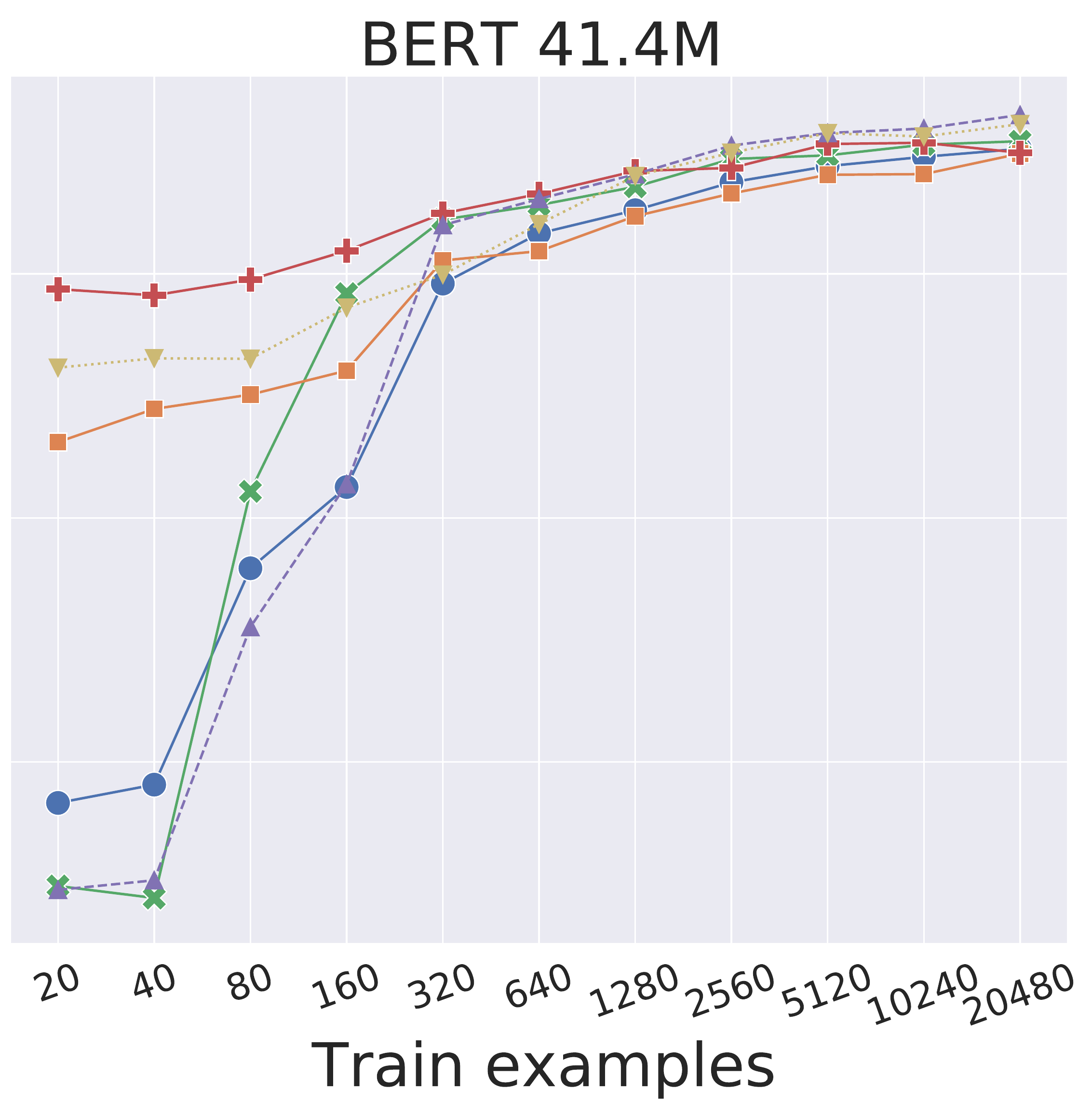}
        \caption{Yahoo-Q}
    \end{subfigure}
    
    \begin{subfigure}[t]{0.45\columnwidth}
        \centering
        \includegraphics[height=1.3in]{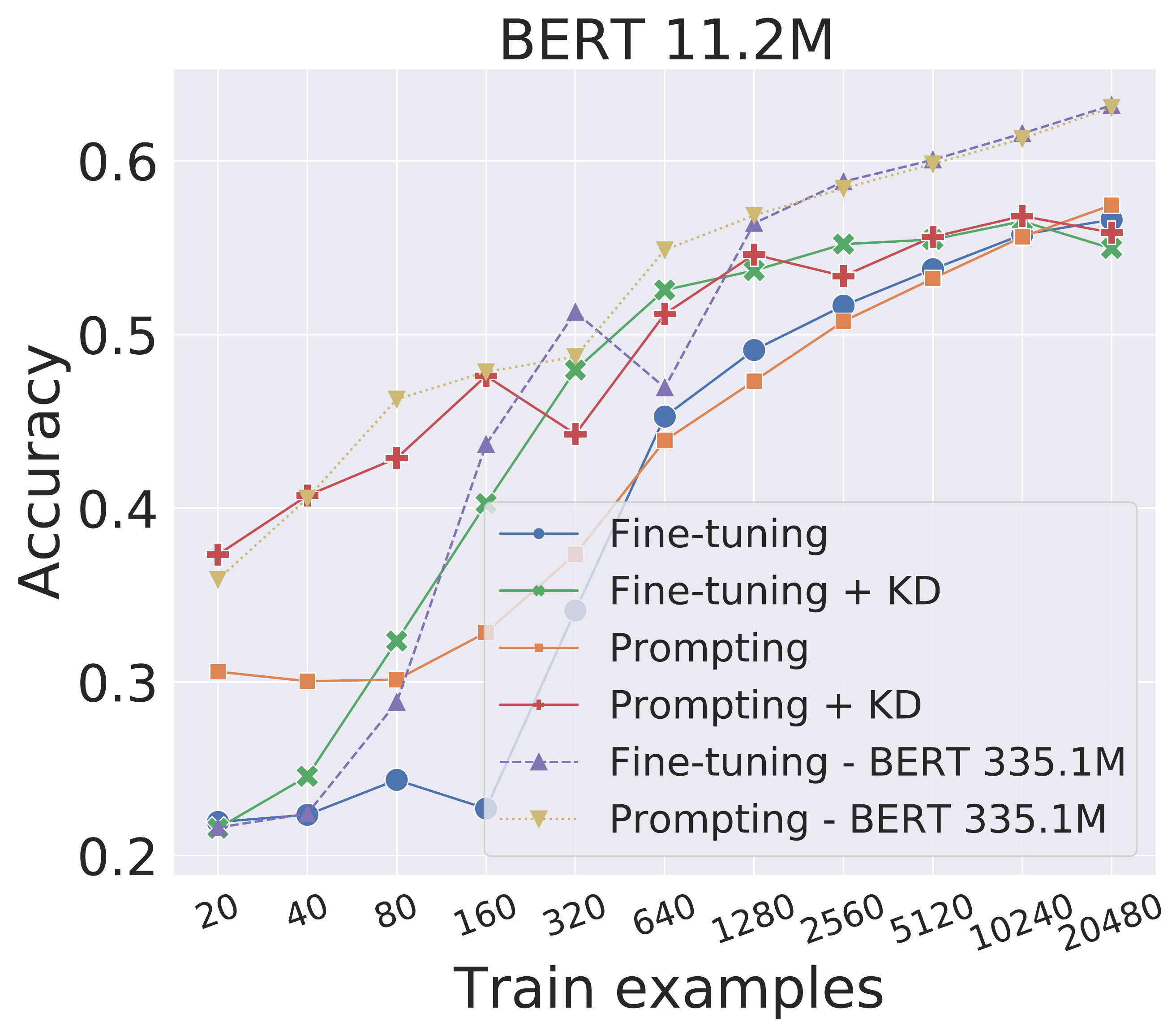}
        \caption{Yelp-Full}
    \end{subfigure}
    ~
    \begin{subfigure}[t]{0.45\columnwidth}
        \centering
        \includegraphics[height=1.3in]{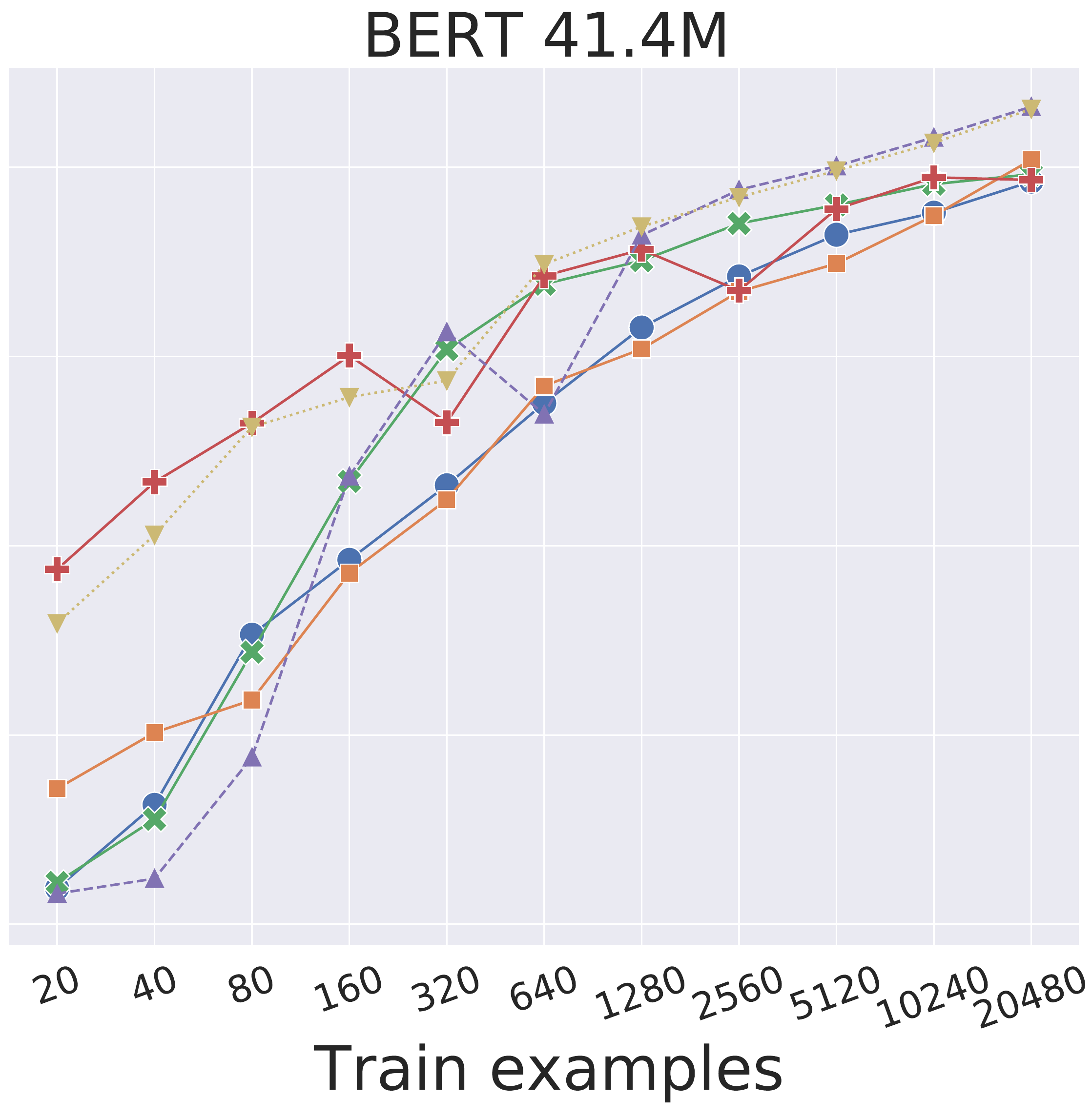}
        \caption{Yelp-Full}
    \end{subfigure}

    \begin{subfigure}[t]{0.45\columnwidth}
        \centering
        \includegraphics[height=1.25in]{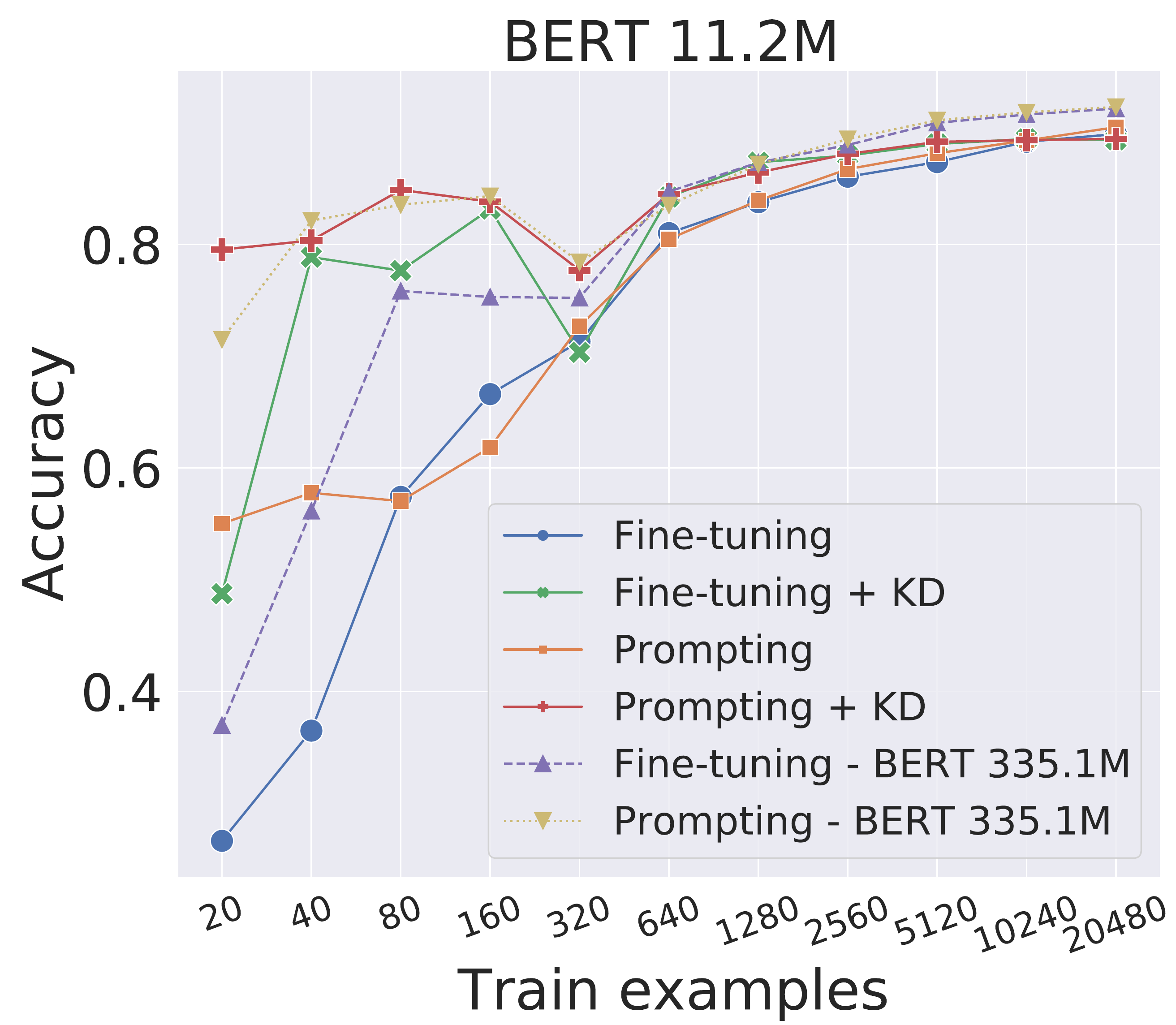}
        \caption{AG News}
    \end{subfigure}
    \begin{subfigure}[t]{0.45\columnwidth}
        \centering
        \includegraphics[height=1.25in]{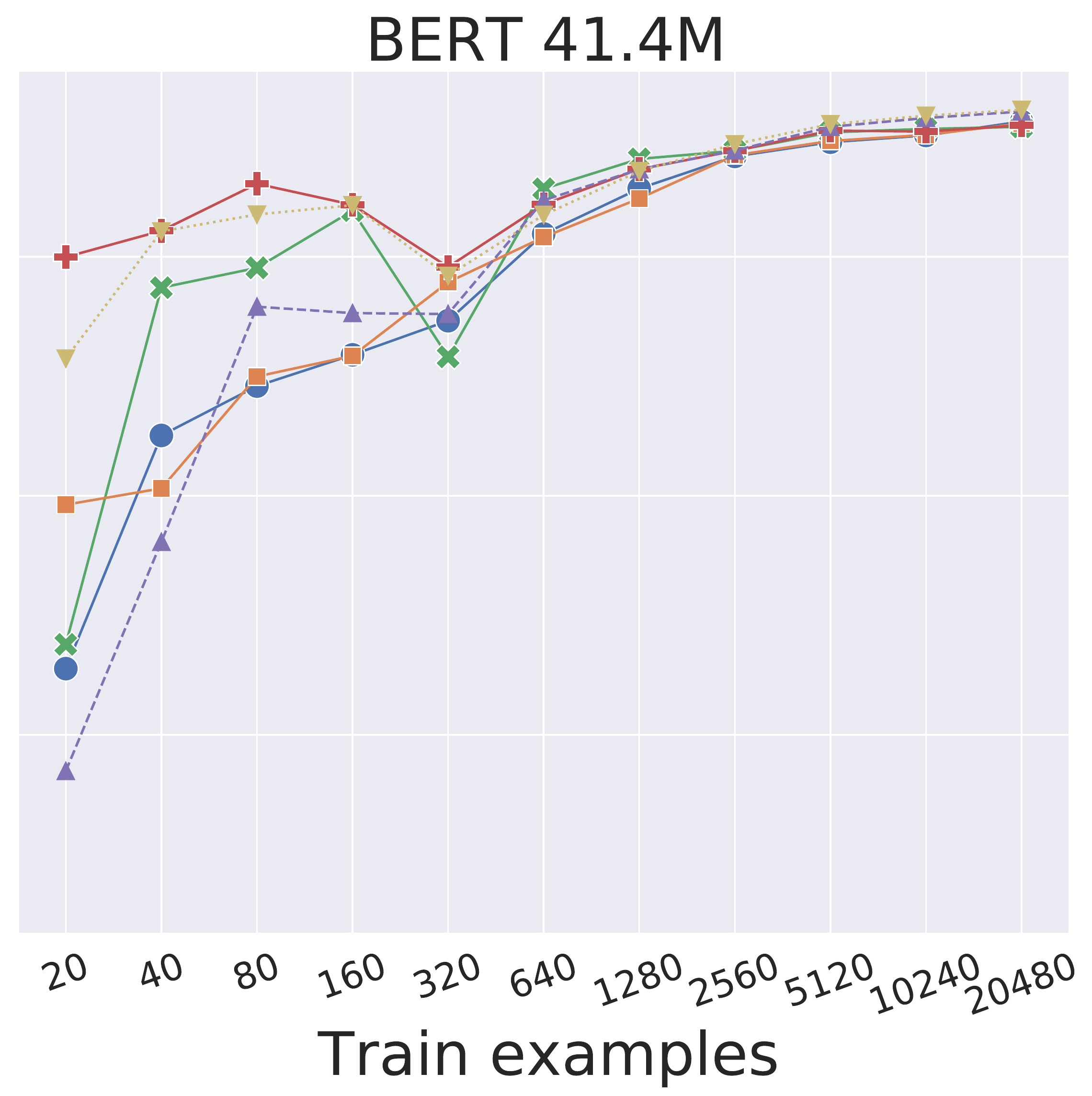}
        \caption{AG News}
    \end{subfigure}
    \caption{Classification accuracy (y-axis) across train set sizes (x-axis) and training procedures (lines) on the BERT 11.2M (mini) and the BERT-41M (medium). To facilitate the comparison we also show the performance of BERT 335M (large), used as teacher in KD.}

    \label{fig:accuracy_yahoo_help_11M}
\end{figure}

\subsection{Effect of Training Procedure}
\label{sec:effect-training}
Fig.~\ref{fig:accuracy_yahoo_help_11M} shows classification accuracy on Yahoo Questions (Yahoo-Q), Yelp-Full and AG News across training procedures, models and train set sizes.\footnote{Please refer to Fig. \ref{fig:accuracy_alltasks} in the Appendix for the complete set of plots for of all model and dataset combinations we considered; the trends are consistent.} 
We identify consistent trends across datasets that can be summarized as follows:
\begin{itemize}
  \setlength\itemsep{0.01em}
\item Prompting (both \emph{P} and \emph{P+KD}) outperforms fine-tuning on small train sets (< 320).
\item Unless the train set is large (> 2560), KD (both \emph{F+KD}, \emph{P+KD}) allows to increase the quality of compact models.
\item \emph{P+KD} leads to effective small models using small train sets (see red lines in Fig.~\ref{fig:accuracy_yahoo_help_11M}). 
\item There is gradually less difference across training procedures as the train set gets larger.
\end{itemize}

\paragraph{Fine-tuning.} In line with ~\citet{le2021many,schick2021exploiting}, we find that fine-tuning performs worse than prompting in few-shot learning (< 320 examples).
There are large accuracy gaps between the smallest and largest train sets (20 and 20K), sometimes starting from random-level performance. 
On very small train sets, using more model parameters is not always beneficial; for instance, BERT-large (335.1M) is worse than BERT 41.4M on Yelp-full for <160 examples.\footnote{An explanation is that smaller models, because of the fewer number of parameters to update, may be less sensitive to model initialization. Another factor that can explain some of the fluctuations in accuracy is that occasionally when the dev set is very small, the best model on it is not necessarily the best on test data. This should not be taken as a limitation of our experiments but rather as evidence of issues which could realistically occur in scenarios with limited data.}
As a result, KD (which uses BERT-large as teacher) does not always improve over fine-tuning. 
However, with more training data, bigger models perform better and KD allows smaller models to fill the gap with the teacher. 
 
\paragraph{Prompting.} Prompting near-always improves the performance over fine-tuning with small train sets (< 320), with or without KD.
Interestingly, smaller LMs trained with prompting are almost always improved with KD (\emph{P+KD}). 
This can lead to having compact models with high accuracy in the low data regime: for instance, BERT with 11M parameters performs comparably with BERT 335.1M for small train sets (<320, Fig.~\ref{fig:accuracy_yahoo_help_11M}), even though it has 30x fewer parameters and 7x faster inference~\citep{turc2019}.
As the train set becomes larger, accuracy increases, 
and there is less difference between fine-tuning and prompting. 

\paragraph{KD.} Combined with either fine-tuning or prompting, KD allows to build effective small models, including with limited train set size. 
Larger student models tend to fill more the gap with the teacher model, presumably due to their wider capacity (see BERT 11M vs BERT 40M).
For both fine-tuning and prompting, KD does not bring improvements when the teacher is exposed to more than 10240 training examples. 
One explanation is that KD is beneficial only if the transfer set is much larger than the train set the teacher was exposed to, whereas we used 10K unlabeled examples for all train set sizes (\textsection\ref{sec:exp-setup}).
Moreover, the gap between the largest model and the smaller ones always decreases 
with more training examples: with a large train set, there is, to begin with, less to gain by learning from a larger model. 

\subsection{Recommendations for Adapting LMs}
\label{sec:recommendations}

\begin{table}[t]
\centering
\footnotesize{
\begin{tabular}{lll} \toprule 

  &         \multicolumn{2}{c}{model size } \\ \midrule
train &   max. acc. diff. = 0.01 & max. acc. diff = 0.05  \\ \midrule
20480                & [335M] \emph{P}, \emph{F}               & [11M] \emph{P}, \emph{F}, \emph{F+KD}          \\
10240                    & [109M] \emph{P+KD}, \emph{F+KD}         & [11M] \emph{P+KD}, \emph{F+KD}    \\
5120                     & [109M] \emph{P+KD}, \emph{F+KD}         & [11M] \emph{P+KD}, \emph{F+KD}          \\
2560                    & [109M] \emph{P+KD}, \emph{F+KD}      & [11M] \emph{P+KD}, \emph{F+KD}          \\
1280                    & [109M] \emph{P+KD}, \emph{F+KD}         & [11M] \emph{P+KD}, \emph{F+KD}          \\
640                      & [109M] \emph{P+KD}, \emph{F+KD}         & [11M] \emph{P+KD}, \emph{F+KD}          \\
320                      & [109M] \emph{F+KD}              & [11M] \emph{F+KD}          \\
160                      & [109M] \emph{P+KD}               & [11M] \emph{P+KD}                  \\
80                       & [109M] \emph{P+KD}                & [11M] \emph{P+KD}                  \\
40                       & [109M] \emph{P+KD}               & [11M] \emph{P+KD}                  \\
20                      & [28M] \emph{P+KD}                 & [11M] \emph{P+KD}                  \\ \midrule
  & \multicolumn{2}{c}{train set size} \\ \midrule

params  &   max. acc. diff. = 0.01 & max. acc. diff = 0.05  \\ \midrule
335M & [10240] \emph{P}                 & [2560] \emph{P}, \emph{F}                   \\
109M & [10240] \emph{P+KD}, \emph{F+KD}        & [1280] \emph{P+KD}, \emph{F+KD}      \\
41M & [5120] \emph{P+KD}               & [1280] \emph{P+KD}, \emph{F+KD}          \\
28M  & [10240] \emph{P+KD}, \emph{F+KD}       & [1280] \emph{P+KD}, \emph{F+KD}        \\
11M  & [10240] \emph{P+KD}, \emph{F+KD}        & [1280] \emph{P+KD}, \emph{F+KD}         \\ \bottomrule 
\end{tabular}}
\caption{Combination of training procedure and smallest train set/model size leading to optimal accuracy (has at most 0.01 or 0.05 accuracy difference from the highest achieved for that train set or model size), for each model/train set size, in at least 3 of the 4 tasks considered. \emph{F}: fine-tuning, \emph{P}: prompting, \emph{F+KD}: fine-tuning followed by KD, \emph{P+KD}: prompting followed by KD.}
\label{table:best}
\end{table}

Here, we explore which combinations of training procedure, train set and model size lead to high performance with low data and compute cost. 
Note that in real-world applications, trading some performance (e.g., accuracy) for efficiency may be acceptable. This is because small differences in offline metrics such as accuracy may not strongly affect the actual efficacy of models in online metrics~\citep{yi2013predictive}. Also, savings that efficient models enable may counterbalance the impact of a slightly worse performance.

We structure the search for the most favorable combinations as follows.
For each model size, we seek for the combination of training procedure and smallest train set size that leads to optimal accuracy. For each train set size, we seek for the combination of training procedure and \emph{smallest model that leads to optimal accuracy}. A combination with optimal accuracy is one that has at most 0.01 or 0.05 accuracy difference from the highest achieved for that train set or model size. 

Table~\ref{table:best} shows the results. We observe that, independently of train set size, we can reduce model size:
with a potential accuracy loss of max 0.05, KD allows us to use the smallest LM considered (11M parameters).
If the train set is small, KD should be combined with prompting.
Train set size can be reduced less safely than model size, though still considerably:
across model sizes, medium-to-large train sets seem to work best, but we do not necessarily need the maximum size considered (20480); we can reduce this further (1280) with some accuracy loss.

\section{Conclusion and Future Work}
We investigated performance-efficiency trade-offs of different training procedures for adapting LMs to text classification tasks.
We considered efficiency both as the compute cost associated with inference and the data cost of labeling training examples.
We found that both prompting and fine-tuning work well to train large LMs on large train sets, but there exist more efficient alternatives to build effective models.
To reduce compute cost, one can prompt or fine-tune compact LMs; if the train set is small, provided the availability of a transfer set, KD from a large model should be applied to obtain a small model. 
To reduce data cost, prompting is recommended, better if combined with KD for smaller models. 
To simultaneously reduce both compute and data cost, \emph{P+KD} is the most efficient training procedure. 

Our results can help NLP practitioners to identify the best strategy to follow on adapting pretrained LMs to text classification tasks based on compute budget and the cost of data collection. Note that in this work we focused on the compute of cost of inference rather than that of training due to its larger impact on real-world applications where models are served frequently and on a large scale. 
However, there are differences in training cost among the procedures we investigated (e.g.,~KD requires training both a teacher and a student model), which will have to be taken into account if facing a limit in train cost budget and resources.

For future work, we want to expand our study to other NLP tasks such as sequence tagging, as well as languages other than English.
Also, in our experiments we used a relatively small teacher model to facilitate experimentation; we expect the performance of \emph{P+KD} to increase if using larger LMs with in-context learning~\citep{brown2020language,alex2021raft}, as opposed to prompt-based fine-tuning.

\section*{Acknowledgements} We thank Diego Marcheggiani and Lluís Màrquez for helpful discussions and feedback about this work, and the anonymous reviewers for their comments.

\bibliography{custom}

\begin{thebibliography}{23}
\expandafter\ifx\csname natexlab\endcsname\relax\def\natexlab#1{#1}\fi

\bibitem[{Alex et~al.(2021)Alex, Lifland, Tunstall, Thakur, Maham, Riedel,
  Hine, Ashurst, Sedille, Carlier et~al.}]{alex2021raft}
Neel Alex, Eli Lifland, Lewis Tunstall, Abhishek Thakur, Pegah Maham, C~Jess
  Riedel, Emmie Hine, Carolyn Ashurst, Paul Sedille, Alexis Carlier, et~al.
  2021.
\newblock {RAFT}: A real-world few-shot text classification benchmark.
\newblock In \emph{Thirty-fifth Conference on Neural Information Processing
  Systems Datasets and Benchmarks Track (Round 2)}.

\bibitem[{Brown et~al.(2020)Brown, Mann, Ryder, Subbiah, Kaplan, Dhariwal,
  Neelakantan, Shyam, Sastry, Askell et~al.}]{brown2020language}
Tom Brown, Benjamin Mann, Nick Ryder, Melanie Subbiah, Jared~D Kaplan, Prafulla
  Dhariwal, Arvind Neelakantan, Pranav Shyam, Girish Sastry, Amanda Askell,
  et~al. 2020.
\newblock Language models are few-shot learners.
\newblock \emph{Advances in neural information processing systems},
  33:1877--1901.

\bibitem[{Dehghani et~al.(2021)Dehghani, Arnab, Beyer, Vaswani, and
  Tay}]{dehghani2021efficiency}
Mostafa Dehghani, Anurag Arnab, Lucas Beyer, Ashish Vaswani, and Yi~Tay. 2021.
\newblock The efficiency misnomer.
\newblock \emph{arXiv preprint arXiv:2110.12894}.

\bibitem[{Devlin et~al.(2019)Devlin, Chang, Lee, and
  Toutanova}]{devlin2019bert}
Jacob Devlin, Ming-Wei Chang, Kenton Lee, and Kristina Toutanova. 2019.
\newblock {BERT}: Pre-training of deep bidirectional transformers for language
  understanding.
\newblock In \emph{Proceedings of the 2019 Conference of the North American
  Chapter of the Association for Computational Linguistics: Human Language
  Technologies, Volume 1 (Long and Short Papers)}, pages 4171--4186.

\bibitem[{Dodge et~al.(2020)Dodge, Ilharco, Schwartz, Farhadi, Hajishirzi, and
  Smith}]{dodge2020fine}
Jesse Dodge, Gabriel Ilharco, Roy Schwartz, Ali Farhadi, Hannaneh Hajishirzi,
  and Noah Smith. 2020.
\newblock Fine-tuning pretrained language models: Weight initializations, data
  orders, and early stopping.
\newblock \emph{arXiv preprint arXiv:2002.06305}.

\bibitem[{Hinton et~al.(2015)Hinton, Vinyals, and Dean}]{hinton2015distilling}
Geoffrey Hinton, Oriol Vinyals, and Jeffrey Dean. 2015.
\newblock Distilling the knowledge in a neural network.
\newblock In \emph{NIPS Deep Learning and Representation Learning Workshop}.

\bibitem[{Hoffmann et~al.(2022)Hoffmann, Borgeaud, Mensch, Buchatskaya, Cai,
  Rutherford, Casas, Hendricks, Welbl, Clark et~al.}]{hoffmann2022training}
Jordan Hoffmann, Sebastian Borgeaud, Arthur Mensch, Elena Buchatskaya, Trevor
  Cai, Eliza Rutherford, Diego de~Las Casas, Lisa~Anne Hendricks, Johannes
  Welbl, Aidan Clark, et~al. 2022.
\newblock Training compute-optimal large language models.
\newblock \emph{arXiv preprint arXiv:2203.15556}.

\bibitem[{Izsak et~al.(2021)Izsak, Berchansky, and Levy}]{izsak2021train}
Peter Izsak, Moshe Berchansky, and Omer Levy. 2021.
\newblock How to train {BERT} with an academic budget.
\newblock In \emph{Proceedings of the 2021 Conference on Empirical Methods in
  Natural Language Processing}, pages 10644--10652.

\bibitem[{Kaplan et~al.(2020)Kaplan, McCandlish, Henighan, Brown, Chess, Child,
  Gray, Radford, Wu, and Amodei}]{kaplan2020scaling}
Jared Kaplan, Sam McCandlish, Tom Henighan, Tom~B Brown, Benjamin Chess, Rewon
  Child, Scott Gray, Alec Radford, Jeffrey Wu, and Dario Amodei. 2020.
\newblock Scaling laws for neural language models.
\newblock \emph{arXiv preprint arXiv:2001.08361}.

\bibitem[{Le~Scao and Rush(2021)}]{le2021many}
Teven Le~Scao and Alexander~M Rush. 2021.
\newblock How many data points is a prompt worth?
\newblock In \emph{Proceedings of the 2021 Conference of the North American
  Chapter of the Association for Computational Linguistics: Human Language
  Technologies}, pages 2627--2636.

\bibitem[{Liu et~al.(2021)Liu, Yuan, Fu, Jiang, Hayashi, and
  Neubig}]{liu2021pre}
Pengfei Liu, Weizhe Yuan, Jinlan Fu, Zhengbao Jiang, Hiroaki Hayashi, and
  Graham Neubig. 2021.
\newblock Pre-train, prompt, and predict: A systematic survey of prompting
  methods in natural language processing.
\newblock \emph{arXiv preprint arXiv:2107.13586}.

\bibitem[{Meng et~al.(2022)Meng, Huang, Zhang, and Han}]{meng2022generating}
Yu~Meng, Jiaxin Huang, Yu~Zhang, and Jiawei Han. 2022.
\newblock Generating training data with language models: Towards zero-shot
  language understanding.
\newblock In \emph{Advances in Neural Information Processing Systems}.

\bibitem[{Radford et~al.(2018)Radford, Narasimhan, Salimans, and
  Sutskever}]{radford2018improving}
Alec Radford, Karthik Narasimhan, Tim Salimans, and Ilya Sutskever. 2018.
\newblock Improving language understanding by generative pre-training.

\bibitem[{Schick and Sch{\"u}tze(2021{\natexlab{a}})}]{schick2021exploiting}
Timo Schick and Hinrich Sch{\"u}tze. 2021{\natexlab{a}}.
\newblock Exploiting cloze-questions for few-shot text classification and
  natural language inference.
\newblock In \emph{Proceedings of the 16th Conference of the European Chapter
  of the Association for Computational Linguistics: Main Volume}, pages
  255--269.

\bibitem[{Schick and Sch{\"u}tze(2021{\natexlab{b}})}]{schick2021generating}
Timo Schick and Hinrich Sch{\"u}tze. 2021{\natexlab{b}}.
\newblock Generating datasets with pretrained language models.
\newblock In \emph{Proceedings of the 2021 Conference on Empirical Methods in
  Natural Language Processing}, pages 6943--6951, Online and Punta Cana,
  Dominican Republic. Association for Computational Linguistics.

\bibitem[{Schick and Sch{\"u}tze(2021{\natexlab{c}})}]{schick2021s}
Timo Schick and Hinrich Sch{\"u}tze. 2021{\natexlab{c}}.
\newblock It’s not just size that matters: Small language models are also
  few-shot learners.
\newblock In \emph{Proceedings of the 2021 Conference of the North American
  Chapter of the Association for Computational Linguistics: Human Language
  Technologies}, pages 2339--2352.

\bibitem[{Schick and Sch{\"u}tze(2021{\natexlab{d}})}]{schick2021true}
Timo Schick and Hinrich Sch{\"u}tze. 2021{\natexlab{d}}.
\newblock True few-shot learning with prompts--a real-world perspective.
\newblock \emph{arXiv preprint arXiv:2111.13440}.

\bibitem[{Strubell et~al.(2019)Strubell, Ganesh, and
  McCallum}]{strubell-etal-2019-energy}
Emma Strubell, Ananya Ganesh, and Andrew McCallum. 2019.
\newblock Energy and policy considerations for deep learning in {NLP}.
\newblock In \emph{Proceedings of the 57th Annual Meeting of the Association
  for Computational Linguistics}, pages 3645--3650, Florence, Italy.
  Association for Computational Linguistics.

\bibitem[{Turc et~al.(2019)Turc, Chang, Lee, and Toutanova}]{turc2019}
Iulia Turc, Ming-Wei Chang, Kenton Lee, and Kristina Toutanova. 2019.
\newblock Well-read students learn better: On the importance of pre-training
  compact models.
\newblock \emph{arXiv preprint arXiv:1908.08962v2}.

\bibitem[{West et~al.(2022)West, Bhagavatula, Hessel, Hwang, Jiang, Le~Bras,
  Lu, Welleck, and Choi}]{west-etal-2022-symbolic}
Peter West, Chandra Bhagavatula, Jack Hessel, Jena Hwang, Liwei Jiang, Ronan
  Le~Bras, Ximing Lu, Sean Welleck, and Yejin Choi. 2022.
\newblock Symbolic knowledge distillation: from general language models to
  commonsense models.
\newblock In \emph{Proceedings of the 2022 Conference of the North American
  Chapter of the Association for Computational Linguistics: Human Language
  Technologies}, pages 4602--4625, Seattle, United States. Association for
  Computational Linguistics.

\bibitem[{Yao et~al.(2022)Yao, Zheng, Yang, and Yang}]{yao2021nlp}
Xingcheng Yao, Yanan Zheng, Xiaocong Yang, and Zhilin Yang. 2022.
\newblock {NLP} from scratch without large-scale pretraining: A simple and
  efficient framework.
\newblock In \emph{International Conference on Machine Learning}, pages
  25438--25451. PMLR.

\bibitem[{Yi et~al.(2013)Yi, Chen, Li, Sett, and Yan}]{yi2013predictive}
Jeonghee Yi, Ye~Chen, Jie Li, Swaraj Sett, and Tak~W Yan. 2013.
\newblock Predictive model performance: Offline and online evaluations.
\newblock In \emph{Proceedings of the 19th ACM SIGKDD international conference
  on Knowledge discovery and data mining}, pages 1294--1302.

\bibitem[{Zhang et~al.(2015)Zhang, Zhao, and LeCun}]{zhang2015character}
Xiang Zhang, Junbo Zhao, and Yann LeCun. 2015.
\newblock Character-level convolutional networks for text classification.
\newblock \emph{Advances in neural information processing systems}, 28.

\end{thebibliography}

\section*{Appendix}
\appendix

\section{Training \& Model development}
\label{sec:appendix-a}

\label{sec:bert-models}

\label{sec:search}

\paragraph{Hyperparameter Search.} We make a set of simplifications to the hyperparameter search process to considerably speed up the process (i.e., not having to run a search for each combination of task, training procedure, model and train set size) while still aiming to a later fair comparison in our experiments. 

\begin{enumerate}
\item For prompting/fine-tuning batch size and learning rate, we run the search only for one task and dataset (Yelp-full), generalizing the optimal configurations to others. 
For the prompting template, we run the search for each task (except for Yelp-polarity, where we can use the identical templates as Yelp-full). In the template search on tasks other than Yelp-full, we set batch size and learning rate based on the Yelp-full search. 
\item We consider only a few combinations of train set and model sizes, generalizing the optimal configurations to analogous setups, based on some ranges. In particular: 20, 320, 2560 examples; BERT-large, BERT-base, BERT-small. Configurations picked for 20 examples are generalized to any value between 20 and 320, etc. BERT-small configurations are generalized to the other two models with close size (BERT-mini, BERT-medium). These choices are based on the assumption that models of close size or trained on train sets of close size should work well with the same hyperparameters configurations.
\item We only run the search for: 1) fine-tuning, and 2) prompting - both without KD. We use the optimal configurations from fine-tuning any time we need to train a classifier from a certain LM, including when training it as a student model with KD.
\end{enumerate}

To run the search, we run the training 4 times, and consider the maximum average dev accuracy (with mimimum standard deviation in case of ties) to establish the optimal configuration. We consider the following values: 

\begin{enumerate}
\item Learning rate: 1e-05, 2e-05, 5e-05;
\item Batch size: 8, 16, 32; for BERT-large, for memory reason we set the batch size to 8 but effectively obtain batch sizes > 8 by modulating the number of steps to accumulate gradients.
\item Prompting template: We use the task-specific templates from~\citet{schick2021exploiting}; 4 for Yelp-full and Yelp-polarity; 6 for other tasks (see Section \ref{sec:templates})
\end{enumerate}

We run grid search for fine-tuning and bayesian search (maximum 18 models) for prompting, due to the bigger set combinations to try for the latter. 

In Table \ref{table:hyperparameters} we report the selected hyperparameters based on the search, jointly with the standard deviation in average dev performance across configurations. The selected hyperparameters for each combination of train set and model size tend to vary, and their choice can be impactful on the achieved accuracy.

\begin{table*}[t] \footnotesize{
\begin{tabular}{lccc} \toprule 
\multicolumn{4}{c}{\textbf{Fine-tuning}} \\ 
& \multicolumn{3}{c}{train set size} \\
parameters 
& 20                        & 320                      & 2560                     \\ \toprule
28M        & \begin{tabular}{l}  Yelp: 2e-05, 8 (std: 0.04)   \end{tabular}   & \begin{tabular}{l} Yelp: 5e-05, 8(std: 0.05)    \end{tabular} & \begin{tabular}{l} Yelp:  1e-05,32 (std: 0.01)    \end{tabular}   \\ \midrule
109.5M     & \begin{tabular}{l} Yelp: 2e-05, 8 (std: 0.03)   \end{tabular}   & \begin{tabular}{l} Yelp:  2e-05, 16 (std: 0.04)   \end{tabular}    & \begin{tabular}{l} Yelp:  1e-05,32 (std: 0.02)   \end{tabular}   \\ \midrule
335.1M     & \begin{tabular}{l}Yelpl:  2e-05, 8 (std: 0.03)  \end{tabular}    & \begin{tabular}{l} Yelp: 2e-05, 8 (std: 0.12)   \\ \end{tabular}   & \begin{tabular}{l}Yelp: 1e-05,32 (std: 0.12)     \end{tabular}      \\ \bottomrule
\multicolumn{4}{c}{\textbf{Prompting}} \\ 
& \multicolumn{3}{c}{train set size} \\
parameters 
& 20                        & 320                      & 2560                     \\ \toprule
28M        & \begin{tabular}{l} Yelp: 1e-05, 32,  0 (std: 0.08)   \\ Yahoo: 0  (std: 0.06) \\ AG news: 3 (std: 0.07) \end{tabular}    & \begin{tabular}{l} Yelp: 5e-05, 32, 0 (std: 0.02)   \\ Yahoo: 2 (std: 0.03) \\ AG news: 2  (std: 0.01)
\end{tabular}   & \begin{tabular}{l}Yelp:  2e-05, 8, 0 (std: 0.01)   \\ Yahoo: 4 (std: 0.01) \\ AG news: 1 (std: 0.01) \end{tabular}    \\ \midrule
109.5M     & \begin{tabular}{l} Yelp: 2e-05, 16, 1 (std: 0.07)   \\ Yahoo: 2 (std: 0.06) \\ AG news: 0 (std: 0.03)  \end{tabular}    & \begin{tabular}{l} Yelp: 1e-05, 16, 0 (std: 0.03)   \\ Yahoo: 5 (std: 0.04) \\ AG news:  3 (std: 0.02) \end{tabular}   & \begin{tabular}{l} Yelp:  2e-05, 16, 3 (std: 0.01)   \\ Yahoo: 1 (std: 0.01) \\ AG news: 1 (std: 0.01) \end{tabular}     \\  \midrule
335.1M     & \begin{tabular}{l} Yelp: 1e-05, 16, 3 (std: 0.07)   \\ Yahoo: 5 (std: 0.07) \\ AG news:  1 (std: 0.07) \end{tabular}     & \begin{tabular}{l} Yelp: 1e-05, 32, 3 (std: 0.03)   \\ Yahoo: 5(std: 0.02)  \\ AG news:  3 (std: 0.01) \end{tabular}    & \begin{tabular}{l} Yelp: 1e-05, 16, 3 (std: 0.01)   \\ Yahoo: 0 (std: 0.01) \\ AG news: 0 (std: 0.01) \end{tabular}    \\ \bottomrule
\end{tabular}}
\caption{Best hyperparameters based on search per task (Yelp-Full: learning rate, batch size, pattern id; other tasks: pattern id only). std = standard deviation of average dev performance across hyperparameters configurations. }
\label{table:hyperparameters}
\end{table*}
\paragraph{Early Stopping.} For all models we use as criterion for early stopping performances on dev data: we stop training when dev accuracy does not grow after 3 epochs (1 epoch when using more than 20K examples). 

\paragraph{} For all tasks, we use a maximum sequence length of 256 tokens.

\begin{figure*}
\begin{center}
\begin{subfigure}[b]{1\textwidth}
\includegraphics[width=16cm]{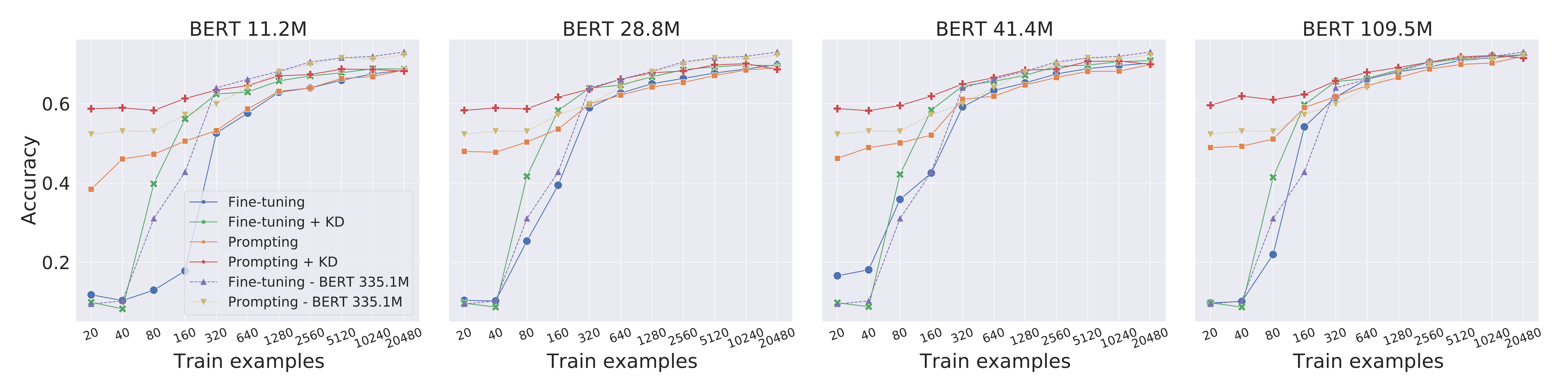}
\caption{Yahoo Questions}
\end{subfigure}
\begin{subfigure}[b]{1\textwidth}
\includegraphics[width=16cm]{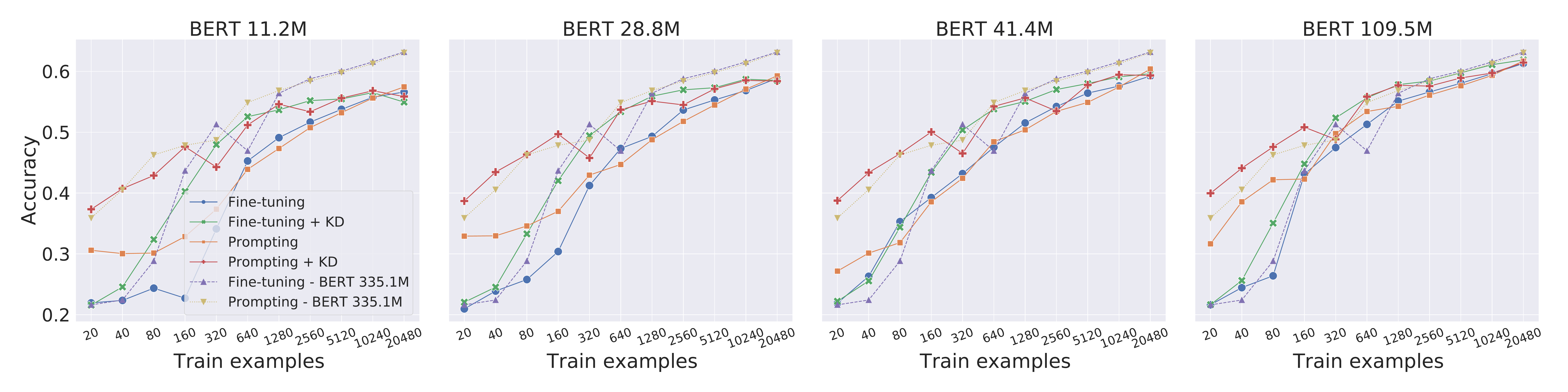}
\caption{Yelp-Full}
\end{subfigure}
\begin{subfigure}[b]{1\textwidth}
\includegraphics[width=16cm]{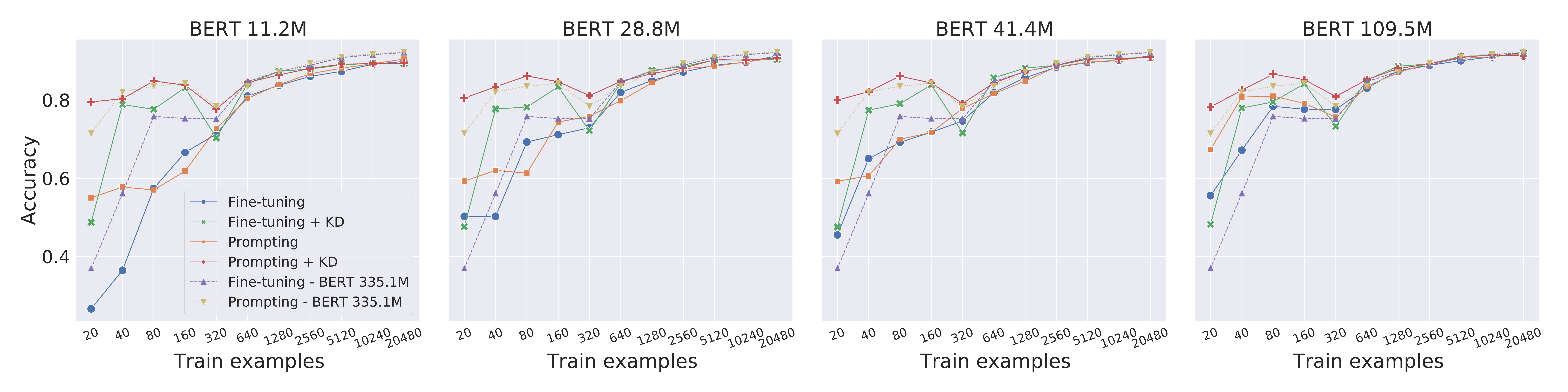}
\caption{AG News}
\end{subfigure}
\begin{subfigure}[b]{1\textwidth}
\includegraphics[width=16cm]{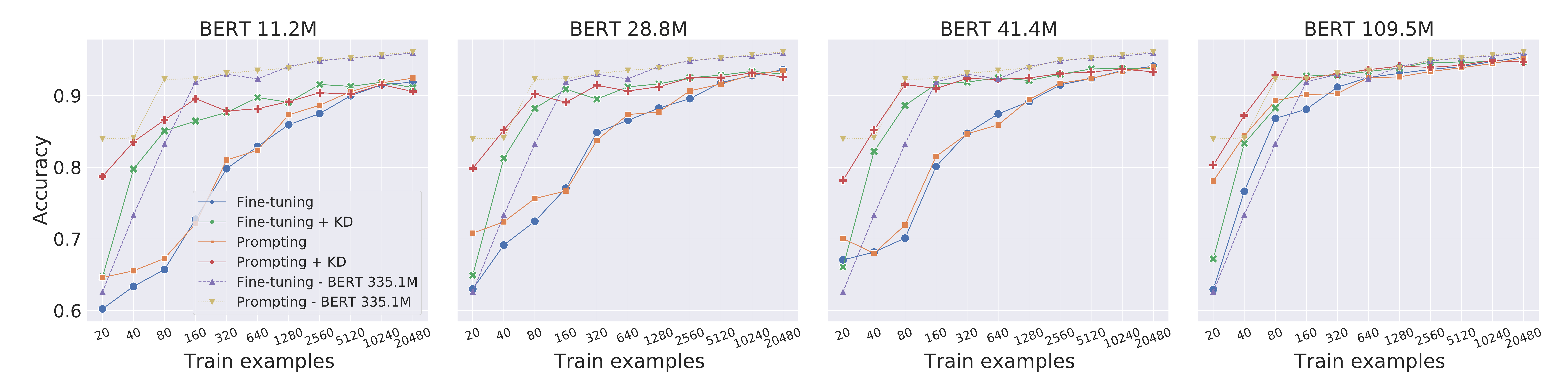}
\caption{Yelp-Polarity}
\end{subfigure}
\end{center}
\caption{Classification accuracy (y-axis) across train set sizes (x-axis), model sizes (plots), and training procedure (lines), for AG News ad Yelp-Polarity. To facilitate the comparison, for each model, we show the performance of BERT-large (335M parameters), used as teachers for KD, in the same plot.}
\label{fig:accuracy_alltasks}
\end{figure*}

\section{Prompting templates}

\label{sec:templates}
\subsection{Yelp-Full \& Yelp-Polarity}

\textbf{Input}:   \\
0) [review] + \textit{It was [MASK] .} \\
1) [review] + \textit{. All in all, it was [MASK] . }\\
2) [review] + \textit{Just [MASK]!} \\
3) [review] + \textit{In summary, the restaurant is [MASK] .} \\
\textbf{Output}:  \\
Yelp-Full: \textit{terrible, bad, okay, good, great} $\rightarrow$ 1-5 \\
Yelp-Polarity: \textit{bad, good} $\rightarrow$ negative, positive. \\

\subsection{Yahoo Questions}

\textbf{Input}: (question, answer)  \\
0) \textit{[MASK] :} + [question] + [answer] \\
1) \textit{[MASK]} \textit{Question:} + [question] + [answer] \\
2) [question] + \textit{([MASK])} + [answer]  \\
3) [question] + [answer] + \textit{([MASK])} \\
4) \textit{[Question: [MASK] ]} + [question] + [answer] \\
5) \textit{[MASK] -} + [question] + [answer] \\
\textbf{Output}:  \\
\textit{Society, Science, Health, Education, Computer, Sports, Business, Entertainment, Relationships, Politics}

\subsection{AG News}

\textbf{Input}: (headline, text)  \\
0) \textit{[MASK] :} + [headline] + [text] \\
1) \textit{[MASK]} \textit{News:} + [headline] + [text] \\
2) [headline] + \textit{([MASK])} + [text]  \\
3) [headline] + [text] + \textit{([MASK])} \\
4) \textit{[News: [MASK] ]} + [headline] + [text] \\
5) \textit{[MASK] -} + [headline] + [text] \\
\textbf{Output}:  \\
\textit{World, Sports, Business, Tech}

\begin{figure*}
\begin{center}
\begin{subfigure}[b]{0.95\textwidth}
\includegraphics[width=14cm]{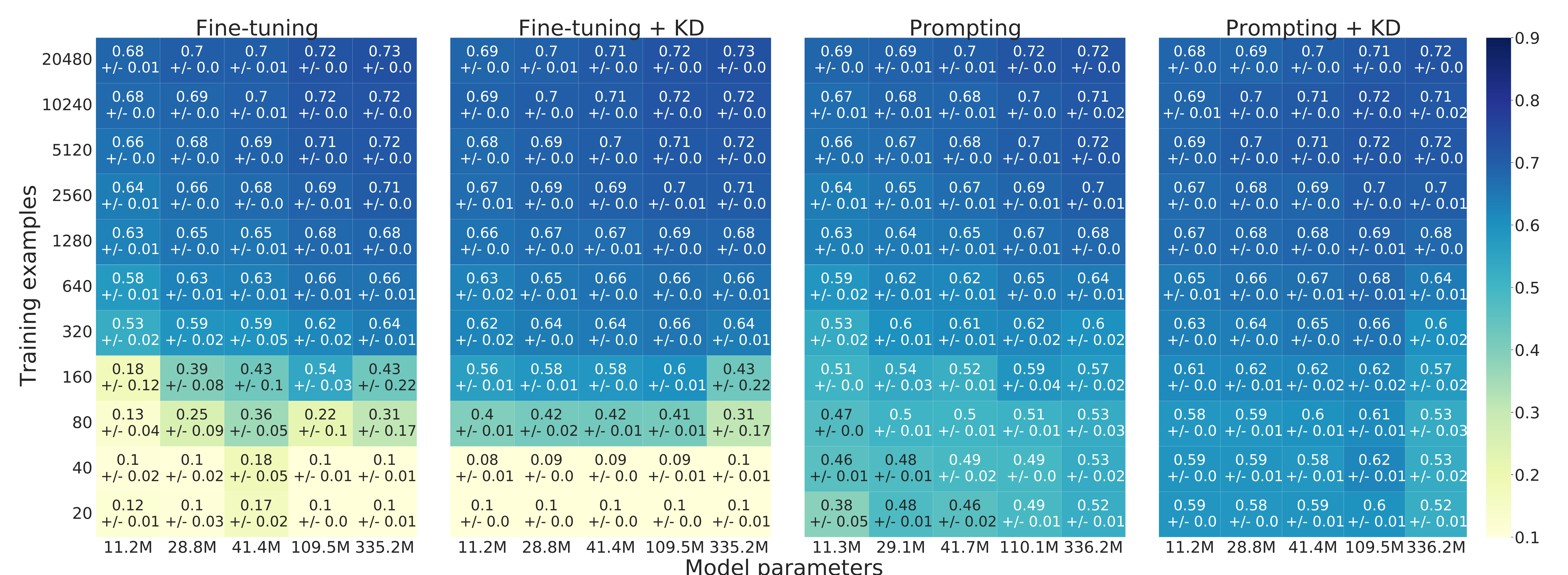}
\caption{Yahoo Questions}
\end{subfigure}
\begin{subfigure}[b]{0.95\textwidth}
\includegraphics[width=14cm]{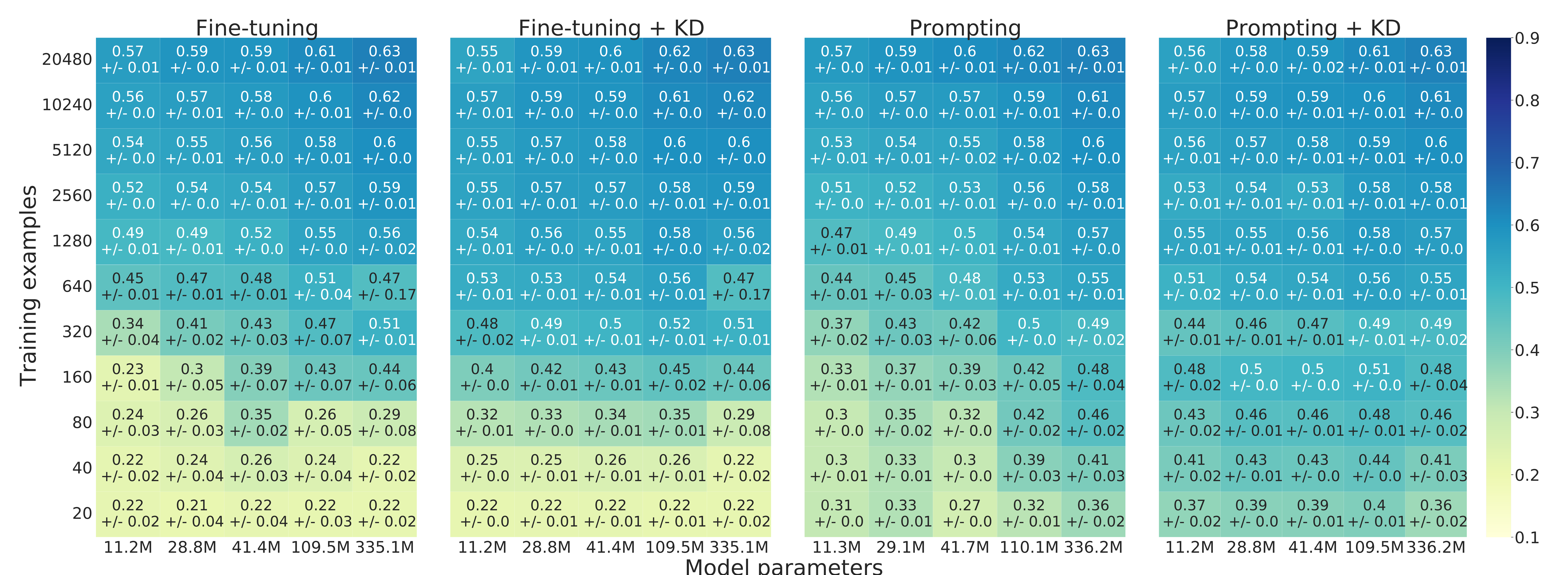}
\caption{Yelp-Full}
\end{subfigure}
\begin{subfigure}[b]{0.95\textwidth}
\includegraphics[width=14cm]{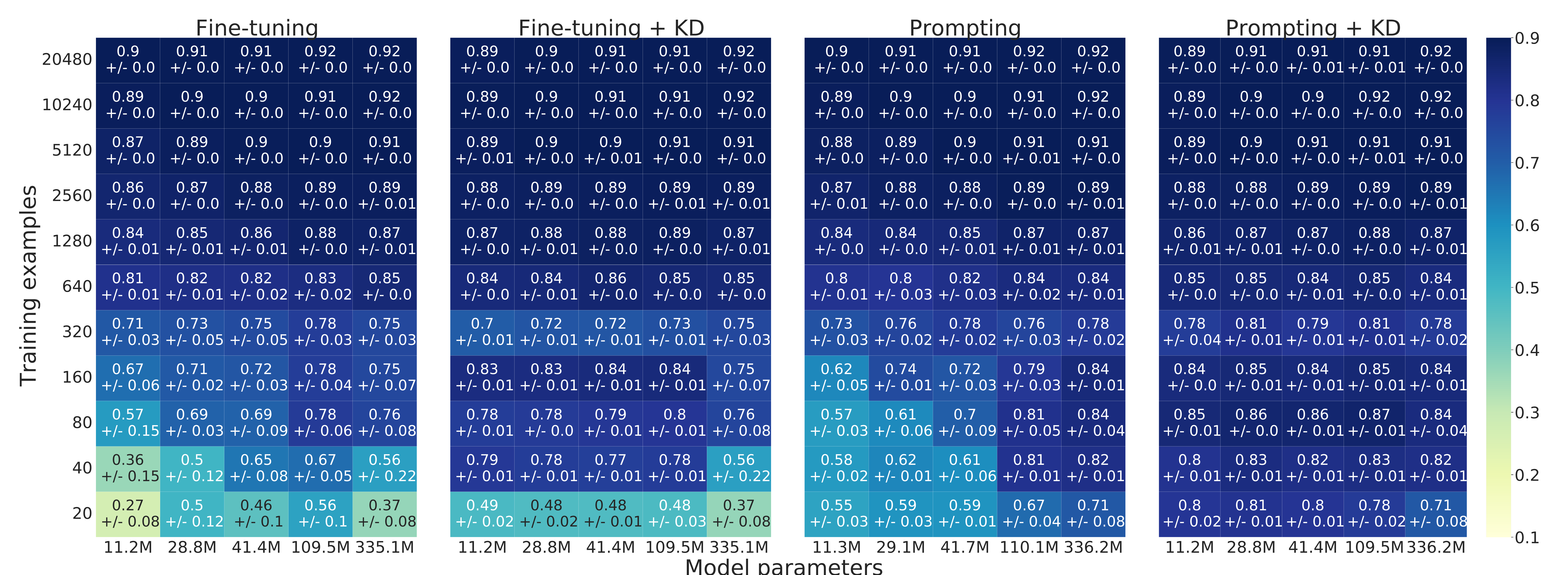}
\caption{AG News}
\end{subfigure}
\begin{subfigure}[b]{0.95\textwidth}
\includegraphics[width=14cm]{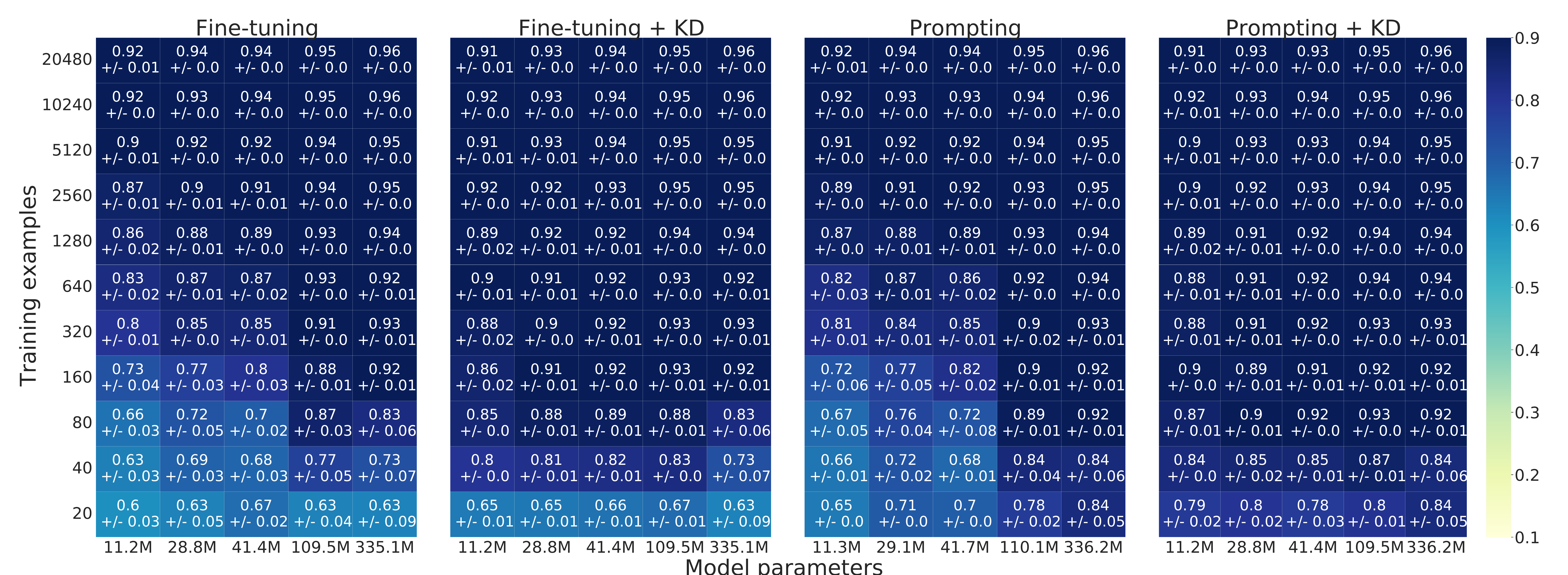}
\caption{Yelp-Polarity}
\end{subfigure}
\end{center}
\caption{For each task, heatmap of mean test classification accuracy (with standard deviation) across models with different number of parameters (x-axis) and train set sizes (y-axis) over 4 training runs.}
\label{fig:accuracy_mean}
\end{figure*}

\end{document}